\documentclass[10pt,twocolumn,letterpaper]{article}

\usepackage[pagenumbers]{cvpr} 

\usepackage{graphicx}
\usepackage{amsmath}
\usepackage{amssymb}
\usepackage{booktabs}

\usepackage{multirow}

\usepackage[pagebackref,breaklinks,colorlinks]{hyperref}

\usepackage[capitalize]{cleveref}
\crefname{section}{Sec.}{Secs.}
\Crefname{section}{Section}{Sections}
\Crefname{table}{Table}{Tables}
\crefname{table}{Tab.}{Tabs.}

\begin{document}

\title{OSIC: A New One-Stage Image Captioner Coined}

\author{Bo~Wang$^{1}$, Zhao~Zhang$^{1,*}$, Mingbo~Zhao$^2$, Xiaojie~Jin$^3$, Mingliang~Xu$^4$, Meng~Wang$^1$
		\vspace{2mm} \\
	$^{1}$ Hefei University of Technology, Hefei, China\\
	$^{2}$ Donghua University, Shanghai, China\\
	$^{3}$ Bytedance Research, USA\\
	$^{4}$ Zhengzhou University, Zhengzhou, China
}
\maketitle

\begin{abstract}
Mainstream image caption models are usually two-stage captioners, i.e., calculating object features by a pre-trained detector and feeding them into a language model to generate text descriptions. However, such an operation will cause a task-based information gap to decrease the performance, because the object features in detection task are suboptimal representations and cannot provide all necessary information for subsequent text generation. Besides, object features are usually represented by the last layer features that lose the local details of images. In this paper, we propose a novel One-Stage Image Captioner (OSIC) with dynamic multi-sight learning, which directly transforms input images into descriptive sentences in one stage. As a result, the task-based information gap is addressed. To obtain rich features, we use the Swin Transformer to calculate multi-level features, and then feed them into a novel dynamic multi-sight embedding module to exploit both the global structure and local texture of input images. To enhance the global modeling of the encoder for caption, we propose a new dual-dimensional refining module to non-locally model the interaction of the embedded features. Finally, OSIC can obtain rich and useful information to improve the image caption task. Extensive comparisons on the benchmark MS-COCO dataset verified the superior performance of our method.

\end{abstract}

\section{Introduction}
\label{sec:intro}
\begin{figure}[t]
	\makeatletter
	\makeatother
	\centering
	\includegraphics[width=\linewidth]{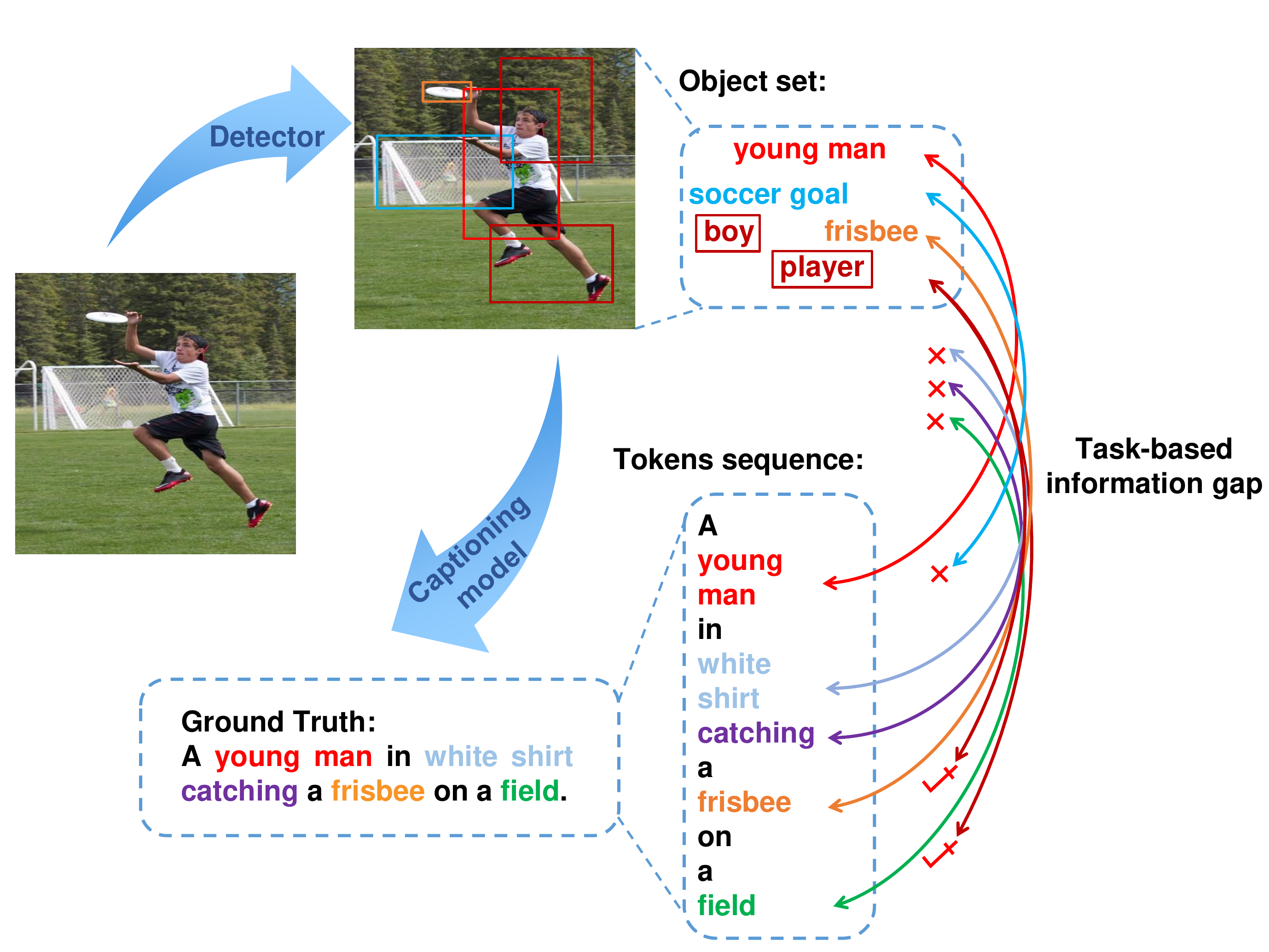}
	\caption{Illustration of the task-based information gap. In the left part, the obtained object features of input image by a fixed detector are inferred to generate the descriptive texts by a language model. Where the red, orange, blue marked young man, frisbee and soccer goal are detected by the pre-trained detector. However, the detection-based object information may be unsuitable for the image caption task. For example, the scene “\emph{field}”, object relation “\emph{catching}” and global detail “\emph{white shirt}” are ignored by the detector, without adaption to the descriptive contents of image. The detected soccer goal is noisy information for the caption task. The boy and player marked in dark red are redundant information for the subject in sentence. The detected object features are mutually independent, which losses the visual semantic connection.}
	\label{fig:task_based_information_gap}
	\vspace{-3mm}
\end{figure}
Image caption, as a multimodal data analysis method, studies the task of automatically generating the natural language sentences to describe the contents of input images. Inspired by the end-to-end model for neural machine translation~\cite{Sutskever2014SequenceTS}, encoder-decoder architecture is the most widely-used framework for image caption~\cite{Karpathy2017DeepVA, Cornia2020MeshedMemoryTF}, which encodes the input image into the intermediate representa-tion via a vision computing model, followed by a natural-language processing model to generate the descriptive texts. As a result, it is crucial to obtain accurate and effective feature representations containing all required information for the language model. Initially, researchers compress the image into the fixed-length vector features~\cite{Vinyals2015ShowAT} as visual features. To enrich the compact expressions, grid features~\cite{Zhang2021RSTNetCW} are generated by the convolutional neural networks (CNN) to embed more visual information. More recently, compared with the grid features, two-stage image caption models using object-based visual representation~\cite{Anderson2018BottomUpAT} have made great progress by recognizing and encoding the salient region-level features.

It is noteworthy that the region-level features obtained by the pre-trained detector (e.g., Faster R-CNN~\cite{Ren2015FasterRT}) focus on the detection task, which means it can hardly provide all necessary descriptive information for the caption task~\cite{Kuo2022BeyondAP}, due to the large difference between the two tasks. In other words, the visual encoder in the first stage of the two-stage captioner is optimized based on the detection tags instead of caption annotation, which may cause information mismatch in feature embedding. We call this mismatch between detection-centric features and caption-guided features \emph{task-based information gap}. Note that this gap will limit the model to obtain a globally optimal solution, and results in two major issues. Firstly, region-level features can hardly present all necessary descriptive objects in the target caption. For example, misdetection, insufficient detection (e.g., the image scene “\emph{field}” and the object relation “\emph{catching}” are not detected) or redundant object information (e.g., “\emph{soccer goal}”, “\emph{boy}” and “\emph{player}”) are produced, as illustrated in Figure~\ref{fig:task_based_information_gap}. And more to the point, these object features are represented independently~\cite{Hu2018RelationNF}, without visual semantic connection with each other. However, as a text sequence, the descriptive sentence will have clear semantic order assigned for each word~\cite{Liu2017MATAM}. Secondly, object information is usually presented by the deepest pooling features~\cite{Ren2015FasterRT, Dai2016RFCNOD}, which may lose the local details of the input image. For example, the young man’s detail “\emph{white shirt}” is missing, and the inadequate descriptions will also decrease the captioning performance.

In this paper, we integrate the intermediate feature representation and caption generation into a unified model, in order to obtain a global optimal solution by a one-stage captioner. Few existing works have discussed the idea of one-stage caption, such as~\cite{Wang2022EndtoEndTB, Fang2022InjectingSC, Zeng2022ProgressiveTP}, which extracts the concept representations based on the outputs of vision transformer (ViT)~\cite{Dosovitskiy2021AnII} or Swin Transformer~\cite{Liu2021SwinTH}, and then embeds the visual features for image captioning. However, existing one-stage caption models just stated that object features cannot represent all the needed information for caption, but they neither show the reasons nor provide analysis. In addition, these methods use the embedded features of fixed sight, so they cannot flexibly capture effective visual information of different sizes, and discover the visual relationships of different distances. While we embed the visual features using the multi-level output of Swin Transformer and calculate the dynamic correlation between the embedded features of different sights, so as to dynamically embed multi-sight features to obtain interconnected visual representation. We also consider to refine the features to generate richer and more accurate descriptive texts. Overall, the main contributions of the paper are summarized as follows:

\begin{itemize}
	\item In this paper, we first clearly define the task-based information gap in image caption models as the presentation mismatch between detection-centric features and caption-guided features. We propose a novel one-stage image captioner (OSIC), which is dedicated to optimizing the captioning architecture and feature representations, in order to reduce the task-based information gap. Technically, a new one-stage captioner with dynamic multi-sight learning encoder is contributed to solve the existed shortages by refining multi-sight features to infer accurate descriptive captions. Experiments on benchmark dataset MS-COCO~\cite{Lin2014MicrosoftCC} with Karpathy's splits~\cite{Karpathy2017DeepVA} showed the state-of-the-art performance of our captioner.
	\vspace{-1mm}
	
	\item Technically, we propose a new idea of dynamic multi-sight embedding (DMSE) to adaptively capture and fuse the global structure in large sight and local texture in small sight. Specifically, it computes the correlation coefficient of the embedded features in different sights, based on the long and short distance dependences of the Swin Transformer. As such, the proposed model can embed the relevant information in different sights into the intermediate representation dynamically.
		\vspace{-1mm}
	
	\item In order to improve the global interaction of the Swin Transformer, we propose a cascaded dual-dimensional refining (DDR) module that non-locally enables the information interaction of features in spatial and channel dimensions in a cascaded mode, so that the global representation ability of encoder can be fully enhanced.
\end{itemize}

\section{Related Work}
\label{sec:related}
We briefly review the existing image captioning methods in terms of the connections between vision and language.

\subsection{Pixel Level-based Representation}
\label{sec:related_a}
\textbf{Fixed length vector-based methods.} In the early stage, simple image encoders, such as VGG-16 and Inception network~\cite{Szegedy2015GoingDW}, are used to encode the input image into a feature vector of fixed length as the global representation. Captions are then generated by a long short-term memory (LSTM)~\cite{Hochreiter1997LongSM} network or with attention mechanism. The major issue of using a fixed-length vector as feature representation is that the information from the input image is heavily compressed and mixed. As a result, salient regions may be fused, and spatial information may be discarded.

\textbf{Grid features-based methods.} Inspired by the success of CNN in visual feature extraction for image classification, ResNet101~\cite{He2016DeepRL} has been used without the last pooling and softmax operation to generate grid features for image caption. For example, pre-trained ResNet101 is used as an encoder to extract grid features from given images, which are then fed into a language decoder~\cite{Gan2017SemanticCN}, e.g., LSTM or Transformer~\cite{vaswani2017attention}, to infer target words~\cite{Gao2022ImprovingIC} and to attach importance on salient pixels~\cite{you2016image}.
\begin{figure*}[t]
	\makeatletter
	\makeatother
	\centering
	\includegraphics[width=0.95\linewidth]{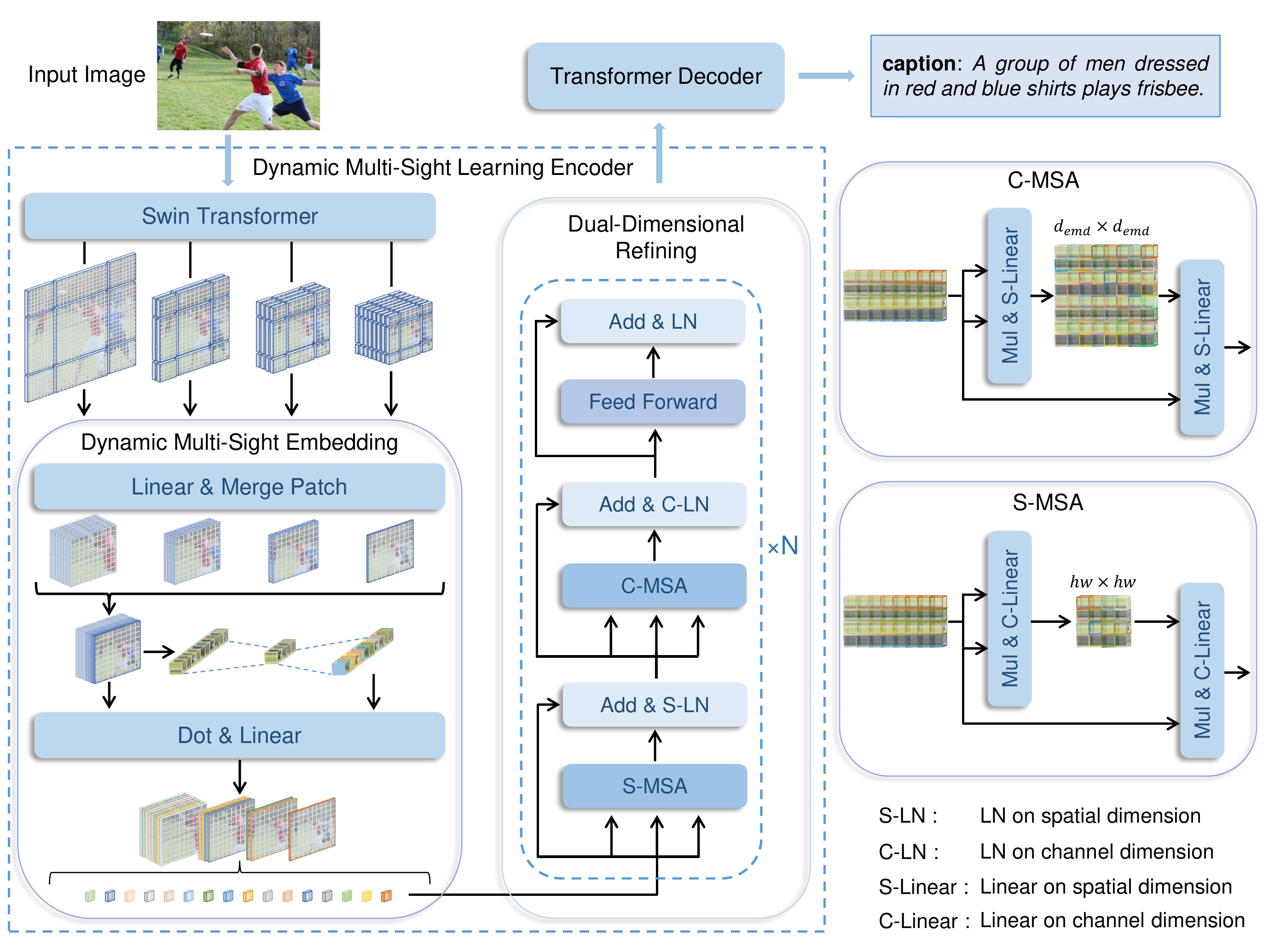}
		\vspace{-2mm}
	\caption{Overview of our OSIC framework that transforms the input image into descriptive sentence by one stage learning. The pipeline of our method includes two main components, i.e., dynamic multi-sight learning encoder and standard Transformer decoder. The dynamic multi-sight learning encoder is built based on the Swin Transformer equipped with dynamic multi-sight embedding (DMSE) and dual-dimensional refining (DRR) modules. Specifically, Swin Transformer is firstly used to obtain grid features from input image, and then the DMSE module calculates the correlation coefficient over the grid features of different sights to dynamically embed the multi-sight features. DRR module further non-locally models the information dependence in spatial and channel dimensions, respectively. Finally, the Transformer decoder generates the captions conditioned on the feature sequence from the DDR module.}
	\label{fig:OSIC}
	\vspace{-3mm}
\end{figure*}
Recently, Vision Transformer (ViT)~\cite{Dosovitskiy2021AnII} and its extension, i.e., Swin Transformer~\cite{Liu2021SwinTH}, are used to extract the grid representation to build one-stage image caption models~\cite{Wang2022EndtoEndTB, Fang2022InjectingSC, Zeng2022ProgressiveTP}. For example, Pure Transformer (PureT)~\cite{Wang2022EndtoEndTB} adopts the Swin Transformer as the backbone of the encoder, further computes the relationships between grid features, and uses the pooled global features as global features for image caption. To consider the visual presentation of semantic concepts, Vision Transformer-based image Caption model (VitCAP)~\cite{Fang2022InjectingSC} introduces the ViT-based token network to predict the semantic classification for captioning. However, ViTCAP still introduces other prior knowledge (i.e., multi-label classification as the concept information) to optimize the models. After that, Progressive Tree-Structured prototype Network (PTSN)~\cite{Zeng2022ProgressiveTP} injects the tree-structured prototypes into grid features by hierarchical clustering to obtain semantic-enhanced visual features. This means that ViT and Swin Transformer provide a potential possibility to reduce the learning gap between visual information and text. However, existing one-stage captioners embed the visual features based on the outputs with fixed sight, without adaption to the feature embedding of different distances.

\subsection{Regional Level-based Representation}
	\vspace{-2mm}
\label{sec:related_a}
Compared to the visual representation of grid features, regional feature-based methods extract the object-level information by pre-trained detector~\cite{Anderson2018BottomUpAT}. The salient objects in a given image can be recognized and embedded via a set of feature vectors, which greatly reduces the difficulty of visual semantic embedding and improves the performance of image caption~\cite{Song2021DirectionRT, Luo2021DualLevelCT}. For example, up-down model~\cite{Anderson2018BottomUpAT} encodes the input image with a set of objects (i.e., RoI-pooled features) detected by a frozen Faster-RCNN~\cite{Ren2015FasterRT} pre-trained on Visual Genome~\cite{Krishna2016VisualGC} for object detection and attribute classification. To further compute the spatial geometry relationships~\cite{Hu2018RelationNF} among detected regions, Object Relation Transformer (ORT)~\cite{Herdade2019ImageCT} explicitly incorporates the relative geometric position and size with the semantic relationships to enrich the embedded features for image captioning. Conditioned on the regions outputted from fixed object detector, $M^2$~\cite{Cornia2020MeshedMemoryTF} infers the descriptive sentence through learning a multi-level representation of the relationship between regions, by using a mesh-like connectivity in the decoder to exploit low- and high-level features. Note that those regional level-based methods learn priori knowledge based on the detection tags, which provide salient information for image caption. However, the task-based information gap between the detection-centric features and the caption-guided features makes these two-stage image caption methods to be suboptimal in solution, which may decrease the captioning performance.
\section{Proposed Method}
\label{sec:formatting}
We introduce the OSIC model in detail, and its framework is shown in Figure~\ref{fig:OSIC}. Clearly, it includes a standard Transformer decoder and a new dynamic multi-sight learning encoder, i.e., Swin Transformer equipped with a newly-designed dynamic multi-sight embedding (DMSE) module and a dual-dimensional refining (DDR) module.

\subsection{Captioning Procedure}
Given an input image $I$, the objective of our OSIC model is to infer and obtain a descriptive sentence $S$ to describe the contents of $I$. Firstly, multi-level grid features $G=\{g_i\}$,($i=1$, $2$, $3$, $4$) are learnt by Swin Transformer~\cite{Liu2021SwinTH} from $I$. Then, a linear embedding layer followed by patch merging further extracts the feature presentation $M$ as the grid features of multi-sights from $G$. Specifically, $M$ is the concatenation of grid features $\{g_i\}$. After that, correlation coefficients $E$ of the grid features of different sights are calculated by DMSE through the average pooling and linear projections to squeeze and dynamically excite the salient features in relevant sight. The output of the DMSE module is tiled into a feature sequence as the input of the DDR module, which consists of $N$ operation layers, performing non-local interaction in spatial and channel dimensions.

Each multi-head DDR attention layer is followed by a position-wise feed-forward network (FFN)~\cite{vaswani2017attention} separately, which is formulated as follows:
\begin{equation}
	FFN(X) = max(0, XW_1 + b_1)W_2 + b_2,
	\label{eq:1}
\end{equation}
where $W_1$, $W_2 \in R^{Hd_v \times d}$ denote the weights of the linear projection functions, $b_1$, $b_2 \in R^{Hd_v \times d}$ denote the projection bias, and $max(\cdot, \cdot)$ is the function to output the maximum value of the input. All those parameters are learnable.

Followed by the position-wise FFN, the refined features are finally decoded by a Transformer decoder to generate the descriptive sentence $S={\{w_i\}}^l$, $l \in L$, where $L$ is the sentence length and $w_i$ is the $i$-th word in the sentence.

\subsection{Dynamic Multi-Sight Embedding (DMSE)}
The grid features of different sights should have different correlation coefficients based on the descriptive contents, so it is inadequate to use only the pooling features in the last layer of CNNs. Simultaneously, simply merging the grid features of multi-sights may cause confusion on the feature embedding. Thus, the DMSE module proposes to calculate the correlation coefficients between the multi-sight features and the descriptive contents. By considering the importance of global optimization for image caption, we obtain a group of learnable coefficients based on the global linear projections of multi-sights. Specifically, we squeeze the grid features in each channel by average pooling to obtain a representative value sequence $V_s$, whose length is equal to the number of feature channels. Then, we connect the sequence $V_s$ with the correlation sequence by two layers of linear connections, so the correlation coefficients of grid features in different sights for subsequent captioning can be generated. Formally, the correlation coefficients $E$ are calculated as
\begin{equation}
	E = L_{c}^{d_{c}}(L_{d_{c}}^{c}((L_{Hd_v}^{1}(M^T))^T)),
	\label{eq:2}
\end{equation}
where $M$ denotes the concatenation of grid features $\{g_i\}$, $(\cdot)^T$ denotes the transpose operation, and $L_i^j (\cdot)$ denotes the linear projection to map a tensor with embedding size $i$ in the last dimension into embedding size $j$.

Then, the output of the DMSE module is obtained by dynamically activating the relevant pixels on $M$, by multiplying the correlation coefficients $E$ by $M$ as follows:
\begin{equation}
	M_e= Norm_l(M\cdot E) + M,
	\label{eq:3}
\end{equation}
where $Norm_l(\cdot)$ denotes the layer normalization, which is followed by a shortcut operation. After that, the outputted features of dynamic multi-sights are fed into the following DRR module for further processing.
\subsection{Dual-Dimensional Refining (DDR)}
Given the embedded features $M_e$ from the DMSE module, we further feed them into the DRR module. To improve the global representation of the encoder, we further refine the features $M_e$ via building non-local information interaction in both spatial and channel dimensions. Each non-local information interaction of the two dimensions is modeled by computing the scaled dot product of them.
\begin{table*}[h]
	\setlength{\abovecaptionskip}{0.05cm}
	\centering
	\caption{\textbf{Performance comparisons on MSCOCO Karpathy test split, where B@$N$, M, R, C and S are short for BLEU@$N$, METEOR, ROUGE-L and CIDEr scores.}}
	\begin{tabular*}{\hsize}{@{\extracolsep{\fill}}l c c c c c c c c c c c c c c}
		\midrule
		\multirow{2}{0.01\textwidth}{Methods}& \multicolumn{7}{c}{\textbf{Cross-Entropy Loss}} & \multicolumn{7}{c}{\textbf{CIDEr Score Optimization}}\\
		\makebox[0.01\textwidth][l] &\makebox[0.01\textwidth][c]{B@1} &\makebox[0.01\textwidth][c]{B@2} &\makebox[0.01\textwidth][c]{B@3} &\makebox[0.01\textwidth][c]{B@4} &\makebox[0.01\textwidth][c]{M} &\makebox[0.01\textwidth][c]{R} &\makebox[0.01\textwidth][c]{C} &\makebox[0.01\textwidth][c]{B@1} &\makebox[0.01\textwidth][c]{B@2} &\makebox[0.01\textwidth][c]{B@3} &\makebox[0.01\textwidth][c]{B@4} &\makebox[0.01\textwidth][c]{M} &\makebox[0.01\textwidth][c]{R} &\makebox[0.01\textwidth][c]{C}\\
		\hline
		LSTM \cite{Vinyals2015ShowAT} &- &- &- &29.6 &25.2 &52.6 &94.0 &- &- &- &31.9 &25.5 &54.3 &106.3\\
		SCST \cite{Rennie2017SelfCriticalST} &- &- &- &30.0 &25.9 &53.4 &99.4 &- &- &- &34.2 &26.7 &55.7 &114.0\\
		LSTM-A \cite{Yao2017BoostingIC} &75.4 &-  &   -  & 35.2 & 26.9 & 55.8 & 108.8 & 78.6 &  -   &  -   & 35.5 & 27.3 & 56.8 & 118.3 \\
		RFNet \cite{Jiang2018RecurrentFN}  & 76.4 & 60.4 & 46.6 & 35.8 & 27.4 & 56.5 & 112.5 & 79.1 & 63.1 & 48.4 & 36.5 & 27.7 & 57.3 & 121.9\\
		Up-Down \cite{Anderson2018BottomUpAT}& 77.2 &   -  &   -  & 36.2 & 27.0 & 56.4 & 113.5 & 79.8 &  -   &  -   & 36.3 & 27.7 & 56.9 & 120.1 \\
		GCN-LSTM \cite{Yao2018ExploringVR} & 77.3 &   -  &   -  & 36.8 & 27.9 & 57.0 & 116.3 & 80.5 &  - &  -& 38.2 & 28.5 & 58.3 & 127.6 \\
		ORT \cite{Herdade2019ImageCT}  & 76.6 &   -  &   -  & 35.5 & 28.0 & 56.6 & 115.4 & 80.5 &   -  &   -  & 38.6 & 28.7 & 58.4 & 128.3 \\
		AoANet \cite{Huang2019AttentionOA}  & 77.4 &   -  &   -  & 37.2 & 28.4 & 57.5 & 119.8 & 80.2 &  -   &  -   & 38.9 & 29.2 & 58.8 & 129.8 \\
		X-LAN\cite{Pan2020XLinearAN} &78.0 &62.3 &48.9 &\textbf{38.2} &28.8 &58.0 & 122.0 &80.8 &65.6 &51.4 &39.5 &29.5 &59.2 &132.0\\
		$M^2$T \cite{Cornia2020MeshedMemoryTF} &- &- &- &- &- &- &- & 80.8 &- &- &39.1 &29.2 &58.6 &131.2 \\
		X-Transformer \cite{Pan2020XLinearAN} &77.3 &61.5 &47.8 &37.0 &28.7 &57.5 &120.0 &80.9 &65.8 &51.5 &39.7 &29.5 &59.1 &132.8 \\
		RSTNet \cite{Pan2020XLinearAN} &- &- &- &- &- &- &- &81.8 &- &- &40.1 &29.8 &59.5 &135.6\\
		GET \cite{Ji2021ImprovingIC} &- &- &- &- &- &- &- &81.5 &- &- &39.5 &29.3 &58.9 &131.6\\
		DRT \cite{Song2021DirectionRT} &- &- &- &- &- &- &- &81.7 &- &- &40.4 &29.5 &59.3 &133.2\\
		$S^2$ Transformer \cite{Zeng2022S2TF} &- &- &- &- &- &- &- &81.1 & & &39.6&29.6 &59.1 &133.5\\
		CBTIC \cite{zhou2022compact} &78.0 &62.2 &48.5 &37.5 &\textbf{29.1} &58.2 &122.6 &81.4 &66.5 &51.9 &39.6 &29.9 &59.1 &\textbf{136.9}\\
		BPTOD \cite{Kuo2022BeyondAP} &- &- &- &- &- &- &- &81.5 &- &- &39.7 &\textbf{30.0} &59.5 &135.9\\
		ViTCAP \cite{Fang2022InjectingSC} &- &- &- &35.7 &28.8 &57.6 &121.8 &- &- &- &40.1 &29.4 &59.4 &133.1\\
		OSIC (Ours) & \textbf{78.5} & \textbf{62.8} & \textbf{49.1} &38.0 &\textbf{29.1} &\textbf{58.3} &\textbf{124.2} &\textbf{81.9} &\textbf{66.8} &\textbf{52.6} &\textbf{40.6} &29.6 &\textbf{59.9} &135.4\\
		\midrule
	\end{tabular*}
	\label{tab_1}
	\vspace{-4mm}
\end{table*}

The layer normalization is operated at the corresponding dimension in which the dependence of pixels is computed. The processing output of the spatial position dimension or channel dimension in the DDR layer is obtained as follows:
\begin{equation}
	M_r^i= Norm_l^i\left(\frac{M_e^Q \cdot \left(M_e^K \right)^T}{\sqrt{d_i}} \cdot M_e^V\right) + M_e,
	\label{eq:4}
\end{equation}
where $i$ denotes the non-local interaction in either spatial or channel dimension, $Norm_l^i(\cdot)$ is the layer normalization operated in $i$-th dimension. Then, the dual-dimensional refining in a parallel way can be formulated as follows:
\begin{equation}
	M_r^{pa}= M_r^{s} + M_r^{c} + M_r,
	\label{eq:5}
\end{equation}
where $M_r^s$ denotes the output from the non-local refining layer of single-spatial dimension, and $M_r^c$ denotes the output of the non-local refining layer of single-channel dimension. Furthermore, the dual-dimensional refining in a cascade way generates the sequential calculation as follows:
\begin{equation}
	M_r^{ca}= Norm_l^c\left(\frac{M_r^{sQ} \cdot \left(M_r^{sK} \right)^T}{\sqrt{d_c}} \cdot M_r^{sV}\right) + M_r^s,
	\label{eq:6}
\end{equation}
where $Norm_l^c(\cdot)$ denotes the layer normalization operated in channel dimension, $M_r^{sQ}$, $M_r^{sK}$, and $M_r^{sV}$ are the linear projection presentations of the outputs from the non-local refining layers of the multi-head self-attention on a spatial dimension, respectively, and $d_c$ is the length of the bottom row vector $M_r^{sK}(Hd_v, :)$ of $M_r^{sK}\in R^{Hd_v \times d_m}$, i.e., $d_m$.

After that, the refined grid features $M_r^i$ are fed into the position-wise FFN and sequentially processed by repeating $N$ times the above operation layer (where $N$ is the number of operation layers. The refined features are finally decoded by a standard Transformer decoder to generate sentences.
\subsection{Objective Function}
We use two objective functions for optimization in the training process, following the common practice of widely used benchmarks. The objective functions consist of the cross-entropy loss for the maximum log-likelihood training and the reinforcement learning loss using the CIDEr score as a reward for self-critical (SC) training~\cite{Rennie2017SelfCriticalST}. At the first training stage, conditioned on the captioning model with parameters $\theta$ and ground truth sentence $y_(1:T)^{\ast}$, the cross-entropy loss is calculated for the optimization as follows:
\begin{equation}
	L_{XE}\left(\theta \right)=-\sum_{t=1}^{T}\log \left(p\left(y_{t}^{\ast}\vert y_{1:T}^{\ast}, I, \theta\right)\right).
	\label{eq:15}
\end{equation}
At the self-critical (SC) training stage, the model can be fine-tuned continually by optimizing the non-differentiable CIDEr-D [] score as the reward of reinforcement learning processing, which is formulated as follows:
\begin{equation}
	\triangledown_{\theta}L_{SC}\left(\theta \right)=-\frac{1}{n}\sum_{t=1}^{n}\left(r\left(y_{1:T}^{i}\right)-b\right)\triangledown_{\theta}\log p_{\theta}\left(y_{1:T}^{i}\right),
	\label{eq:16}
\end{equation}
\noindent where $n$ denotes the beam size, $r(\cdot)$ denotes the CIDEr-D score function, and $b=(\sum_{i}r(y_{1:T}^{i}))/n$ is the greedily decoded score value generated by the current model.
\begin{table*}[h]
	\setlength{\abovecaptionskip}{0.05cm}
	\centering
	\caption{\textbf{Ablation study on the importance of each module. DDR module is further ablated as four modes: refining in spatial/channel dimension, and combining them in parallel/cascade way. The result is obtained using Cross-Entropy Loss training on the test set of MS-COCO Karpathy split, where B@N, M, R, C and S are short for BLEU@N, METEOR, ROUGE-L and CIDEr scores.}}
	\begin{tabular*}{\hsize}{@{\extracolsep{\fill}}c c c c c c c c c c c c c }
		\midrule
		\multirow{2}{0.03\textwidth}{\textbf{Baseline}}& \multirow{2}{0.1\textwidth}{\textbf{Multi-sight embedding}}& \multicolumn{4}{c}{\textbf{Feature refining}} & \multicolumn{7}{c}{ }\\
		\makebox[0.01\textwidth][c] &\makebox[0.01\textwidth][c] &\makebox[0.02\textwidth][c]{Spatial} &\makebox[0.02\textwidth][c]{Channel} &\makebox[0.02\textwidth][c]{Parallel} &\makebox[0.01\textwidth][c]{Cascade} &\makebox[0.01\textwidth][c]{B@1} &\makebox[0.01\textwidth][c]{B@2} &\makebox[0.01\textwidth][c]{B@3} &\makebox[0.01\textwidth][c]{B@4} &\makebox[0.01\textwidth][c]{M} &\makebox[0.01\textwidth][c]{R} &\makebox[0.01\textwidth][c]{C}\\
		\hline
		\checkmark & & & & & &68.8 &53.8 &40.3 &29.4 &23.6 &53.1 &92.2\\
		\checkmark &\checkmark & & & & &77.9 &62.1 &48.2 &37.2 &28.8 &57.8 &122.2\\
		\checkmark & &\checkmark & & & &76.9 &60.9 &47.2 &36.3 &28.5 &57.0 &119.2\\
		\checkmark & & &\checkmark & & &77.2 &61.1 &47.2 &36.2 &28.5 &57.2 &119.3\\
		\checkmark & & &&\checkmark & &77.5 &61.5 &47.6 &36.6 &28.5 &57.4 &119.9\\
		\checkmark & & & & &\checkmark &77.5 &61.6 &47.8 &36.8 &28.6 &57.4	&120.9\\
		\checkmark &\checkmark &\checkmark & & & &78.0 &62.4 &48.6 &37.6	&29.0 &58.0	&123.1\\
		\checkmark &\checkmark & &\checkmark& & &78.2 &62.4 &48.6 &37.5	&28.9 &58.0	&122.6\\
		\checkmark &\checkmark & & &\checkmark& &\textbf{78.6} &\textbf{63.0} &\textbf{49.1} &\textbf{38.0}	&28.9 &58.2	&123.1\\
		\checkmark &\checkmark & & & &\checkmark &78.5 &62.8 &\textbf{49.1} &\textbf{38.0} &\textbf{29.1} &\textbf{58.3} &\textbf{124.2}\\
		\midrule
	\end{tabular*}
	\vspace{-4mm}
	\label{tab_2}
\end{table*}

\section{Experiments}
We evaluate our OSIC method for image caption, along with illustrating the comparison results with related models.

\subsection{Experimental Settings}
\textbf{Dataset MSCOCO}~\cite{Lin2014MicrosoftCC} is the widely-used and competitive benchmark for image caption. This dataset has $123,287$ images annotated with $5$ ground truth captions for each image. To fairly compare each model, we follow the Karpathy’s splits~\cite{Karpathy2017DeepVA} for offline evaluation in this study. This split consists of $113,287$ training images, $5,000$ validation images and $5,000$ testing images. In this study, we resize all images of the MS COCO dataset into $384\times384$ for Swin Transformer. To train the caption model stably and efficiently, the dictionary of descriptive words is built by collecting the words that occur more than $5$ times and then ends up with a vocabulary of $9,487$ words. We convert all the training captions to lowercase in the training process. The sentences longer than 16 words are truncated, and this case in ground truth captions account for lower than $0.1\%$.

\textbf{Evaluation Metrics}. For fair evaluation, we use four widely-used metrics to evaluate the quality of the generated captions of each method, i.e., BLEU-$1/4$~\cite{Papineni2002BleuAM}, CIDEr~\cite{Vedantam2015CIDErCI}, METEOR~\cite{Banerjee2005METEORAA} and ROUGE-L~\cite{Lin2004ROUGEAP}. They are denoted as B-$1/4$, C, M, and R for short. All values of the metrics are reported as percentage (\%), and the larger the better.

\textbf{Implementation Details}. We set the embedding size to $512$ and set the number of multi-heads in all self-attention modules to $8$, which follows the same setting as Transformer network~\cite{vaswani2017attention}. For training, we first train our model under cross-entropy loss for $15$ epochs with a mini-batch size of $8$, and an Adam optimizer whose learning rate is initialized at $4e-4$ and the warmup step is set to $20,000$. The learning rate is decayed $0.1$ times and starts from the $9$-th epoch. We increase the scheduled sampling probability by $0.05$ for every $3$ epochs. After the cross-entropy loss training, we train our model by optimizing the CIDEr score with a self-critical training strategy for another $15$ epochs with an initial learning rate of $4e-5$, which is decayed $0.1$ times every $3$ epochs. For testing, we use the beam search for our model with a beam size of $2$. The default random seed is set to $42$. All experiments are conducted in single NVIDIA RTX$2080$Ti GPU with Pytorch $1.7$ platform.

\subsection{Main Results}
\textbf{Numerical Image Captioning Results.} The numerical results of each method are described in Table~\ref{tab_1}. We show the results on the test set with cross-entropy training firstly, and then the self-critical sequence training (SCST)~\cite{Rennie2017SelfCriticalST} using reinforcement learning fine-tuning. We see clearly that a performance gain with $40.6$ in BLEU-$4$ metirc is achieved by our OSIC model. We first compare our method with the conventional grids-/objects-based methods in the $1$-st to $17$-th rows of Table~\ref{tab_1}. By addressing the problem caused by the task-based information gap, our OSIC model improves the performance by $+1.3\%$ in CIDEr and $+2.5\%$ in BLEU-$4$, and compares favorably with all previous methods across almost all metrics. This proves the effectiveness of our new one-stage image captioner. Then, we compare our method to the end-to-end models, including BPTOD~\cite{Kuo2022BeyondAP} and ViTCAP~\cite{Fang2022InjectingSC} in the $18$-th to $19$-th rows. In spite of additional information are used in these methods (e.g., retrieved text descriptions and image conditioning from pre-trained CLIP~\cite{Radford2021LearningTV} for BPTOD, and multi-label classification for ViTCAP), our OSIC model achieves more performance gains using the dynamic multi-sight learning. Note that our model is only trained on the image-text pairs, without other additional information or annotations, so it avoids the annotation cost of dataset. That is, our dynamic multi-sight learning encoder can clearly benefit the image caption task.
\begin{figure}[h]
	\makeatletter
	\makeatother
	\centering
	\includegraphics[width=\linewidth]{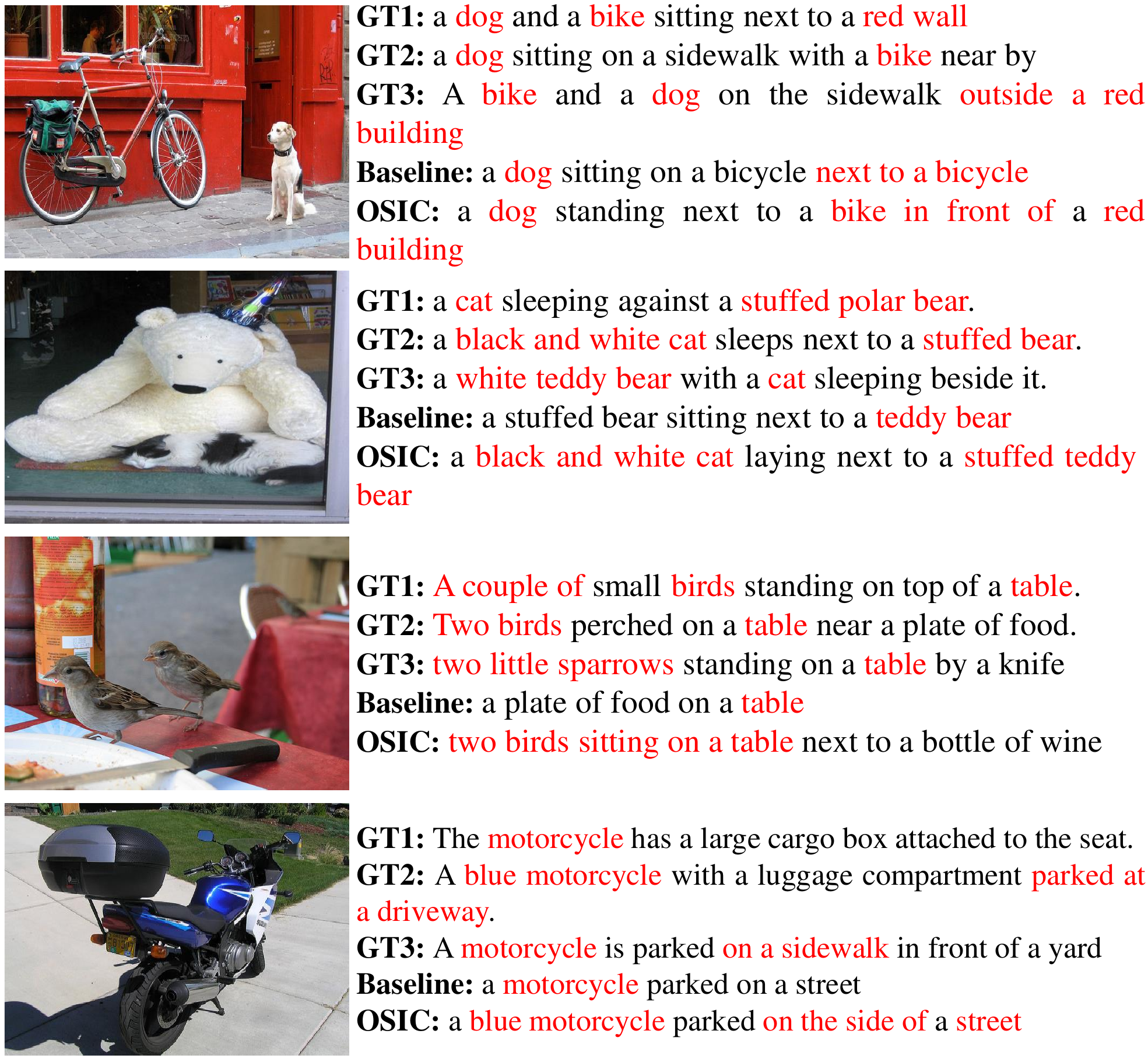}
	\vspace{-6mm}
	\caption{Visualization examples of some generated captions.}
	\label{fig:examples}
	\vspace{-4mm}
\end{figure}

\textbf{Visualizations of Generated Captions.} We show some examples of generated image captions of baseline method and our method in Figure~\ref{fig:examples}. The baseline only uses the output of the standard Swin Transformer as visual information, and is decoded by the standard Transformer decoder. Clearly, our OSIC catches additional details and generate more descriptive captions with accurate semantic relationships, since our OSIC model can dynamically embed and refine the more visual sources of multi-sights as the visual representation for image caption. For example, our model accurately captures the
key scene in top ficture (i.e., ”in front of a red building ”), while the baseline misses it.

\subsection{Ablation Study}
\textbf{1) What is the effect of each module on our performance?} We first explore the importance of each module in OSIC, i.e., the DMSE module, DDR module and Swin Transformer in our dynamic multi-sight encoder. In this study, the model extracting visual features from the last layer of the Swin Transformer, followed by the standard Transformer decoder, is used as the baseline.

\textbf{Evaluation on DMSE module.} We add the DMSE module into the baseline and see the changes in the first and second rows of Table~\ref{tab_2}. Clearly, the DMSE module greatly improves the performance across all metrics over the baseline, which proves the effectiveness of proposed DMSE module.

\textbf{Evaluation on DDR module.} As shown in the 3rd to 6-th row in Table~\ref{tab_2}, we study the effectiveness of the DDR module containing non-local interactions in spatial/channel dimension, or combining them in a parallel/cascade way, respectively. Clearly, refining the features from spatial or channel dimension is of great benefit for our model, with improvements of at least +11.8\% in Bleu 1 and +29.3\% in CIDEr over baseline. The parallel/cascade combination form further improves the performance. Specifically, the cascade way slightly outperforms the parallel way.
\begin{figure}[h]
	\makeatletter
	\makeatother
	\centering
	\includegraphics[width=\linewidth]{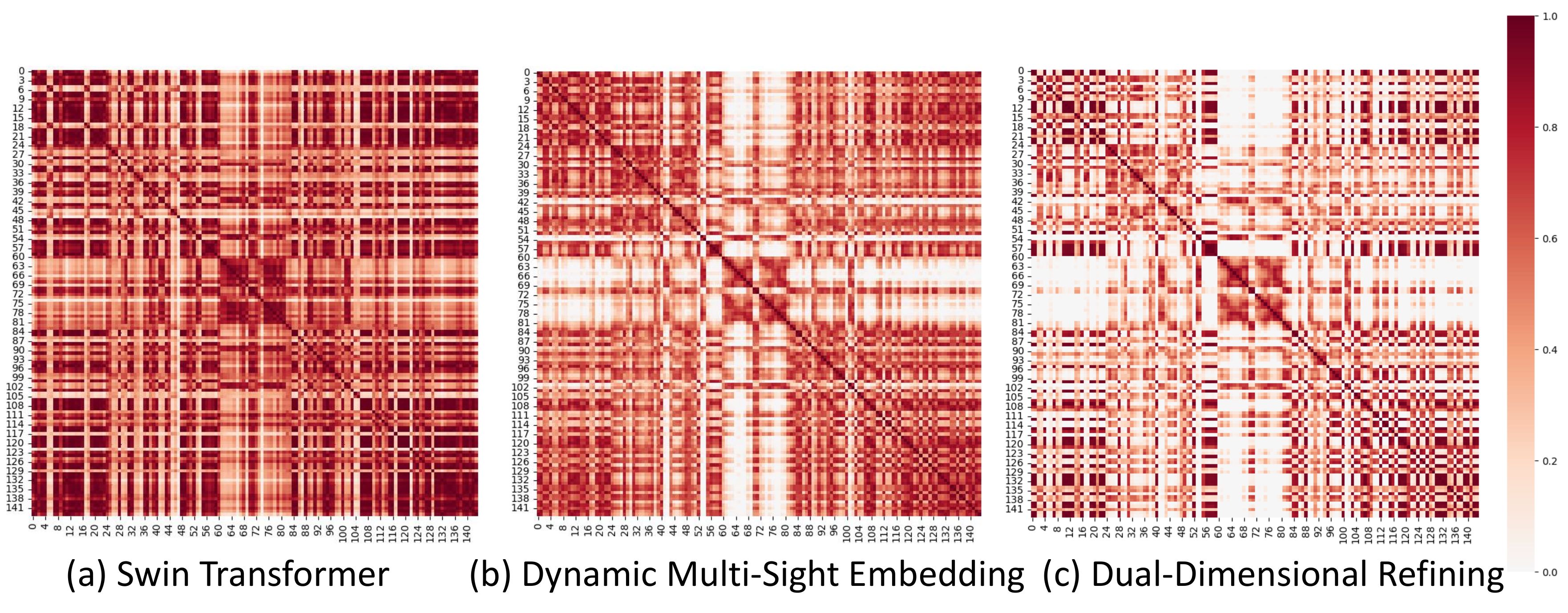}
	\vspace{-6mm}
	\caption{Visualization of the correlations between pixel-level features at different levels by the heatmaps. The degree of red denotes the correlation degree between features. The heavier the color, the closer the pairwise relationship.}
	\label{fig:heatmap_single}
	\vspace{-3mm}
\end{figure}
\begin{table}[h]
	\centering
	\caption{\textbf{Non-local modeling in spatial dimension.}}
	\vspace{-3mm}
	\resizebox{0.475\textwidth}{!}{
		\begin{tabular}{c c c c c c c c}
			\midrule
			Layers &B@1 &B@2 &B@3 &B@4 &M &R &C\\
			\hline
			0 &77.9 &62.1	&48.2 &37.2	&28.8 &57.8 &122.2\\
			1 &78.0 &62.4	&48.6 &37.6	&29.0 &58.0	&123.1\\
			2 &78.5 &62.8	&49.0 &37.9	&29.0 &58.3	&123.7\\
			3 &78.1 &62.5	&48.8 &37.7	&29.0 &58.1	&123.1\\
			4 &78.3 &62.9	&49.3 &38.3	&29.1 &58.4	&123.9\\
			5 &78.3 &62.9	&49.1 &38.1	&28.9 &58.0	&123.3\\
			6 &78.2 &62.6	&48.9 &38.0	&28.9 &58.2	&123.3\\
			\midrule
	\end{tabular}}
	\label{tab_3}
	\vspace{-3mm}
\end{table}
\begin{table}[h]
	\centering
	\caption{\textbf{Non-local modeling in channel dimension.}}
	\vspace{-3mm}
	\resizebox{0.475\textwidth}{!}{
		\begin{tabular}{c c c c c c c c}
			\midrule
			Layers &B@1 &B@2 &B@3 &B@4 &M &R &C\\
			\hline
			0 &77.9 &62.1	&48.2 &37.2 &28.8 &57.8 &122.2\\
			1 &78.2 &62.4	&48.6 &37.5 &29.0 &58.0 &122.6\\
			2 &78.6 &63.0	&49.2 &38.1 &29.0 &58.1 &123.5\\
			3 &78.4 &62.9	&49.2 &38.1 &29.1 &58.3 &123.7\\
			4 &78.2 &62.6	&48.8 &37.7 &29.0 &58.1 &122.8\\
			5 &78.5 &62.8	&49.0 &37.8 &29.0 &58.2 &123.8\\
			6 &78.4 &62.3	&48.3 &37.2 &28.9 &57.9 &122.2\\
			\midrule
	\end{tabular}}
	\label{tab_4}
	\vspace{-4mm}
\end{table}
\begin{table}[h]
	\centering
	\caption{\textbf{Comparison of non-local modeling in parallel way.}}
	\vspace{-3mm}
	\resizebox{0.475\textwidth}{!}{
		\begin{tabular}{c c c c c c c c}
			\midrule
			Layers &B@1 &B@2 &B@3 &B@4 &M &R &C\\
			\hline
			0 &77.9 &62.1 &48.2 &37.2	&28.8 &57.8 &122.2\\
			1 &78.6 &63.0 &49.1 &37.9	&28.9 &58.2 &123.1\\
			2 &78.4 &62.7 &48.9 &37.8	&28.9 &58.1 &123.1\\
			3 &77.9 &62.3 &48.5 &37.6	&28.9 &57.8 &122.4\\
			4 &78.0 &62.4 &48.7 &37.6	&28.8 &58.1 &122.5\\
			5 &78.1 &62.2 &48.3 &37.0	&28.7 &57.7 &121.2\\
			6 &77.5 &61.6 &47.7 &36.6	&28.3 &57.3 &119.8\\
			\midrule
	\end{tabular}}
	\label{tab_5}
	\vspace{-3mm}
\end{table}
\begin{table}[h]
	\centering
	\caption{\textbf{Comparison of non-local modeling in cascade way.}}
	\vspace{-3mm}
	\resizebox{0.475\textwidth}{!}{
		\begin{tabular}{c c c c c c c c}
			\midrule
			Layers &B@1 &B@2 &B@3 &B@4 &M &R &C\\
			\hline
			0 &77.9 &62.1 &48.2 &37.2	&28.8 &57.8 &122.2\\
			1 &78.5 &62.8 &49.1 &38.0 &29.1 &58.3 &124.2\\
			2 &78.3 &62.7 &49.1 &38.0	&29.2 &58.3 &123.9\\
			3 &78.4 &62.8 &49.1 &37.9	&29.1 &58.3 &123.9\\
			4 &78.2 &62.5 &48.7 &37.7	&28.9 &58.0 &122.7\\
			5 &78.0 &62.3 &48.5 &37.5	&29.0 &58.0 &122.9\\
			6 &77.8 &62.0 &48.4 &37.5 &28.9 &58.0 &122.5\\
			\midrule
	\end{tabular}}
	\label{tab_6}
	\vspace{-4mm}
\end{table}

\begin{table*}[h]
	\setlength{\abovecaptionskip}{0.05cm}
	\centering
	\caption{\textbf{Performance comparisons on MSCOCO Karpathy test split, where B@$N$, M, R, C and S are short for BLEU@$N$, METEOR, ROUGE-L and CIDEr scores. All values are reported as percentage (\%).}}
	\begin{tabular*}{\hsize}{@{\extracolsep{\fill}}l c c c c c c c c c c c c c c}
		\midrule
		\multirow{2}{0.01\textwidth}{Models}& \multicolumn{7}{c}{\textbf{Cross-Entropy Loss}} & \multicolumn{7}{c}{\textbf{CIDEr Score Optimization}}\\
		\makebox[0.01\textwidth][l] &\makebox[0.01\textwidth][c]{B@1} &\makebox[0.01\textwidth][c]{B@2} &\makebox[0.01\textwidth][c]{B@3} &\makebox[0.01\textwidth][c]{B@4} &\makebox[0.01\textwidth][c]{M} &\makebox[0.01\textwidth][c]{R} &\makebox[0.01\textwidth][c]{C} &\makebox[0.01\textwidth][c]{B@1} &\makebox[0.01\textwidth][c]{B@2} &\makebox[0.01\textwidth][c]{B@3} &\makebox[0.01\textwidth][c]{B@4} &\makebox[0.01\textwidth][c]{M} &\makebox[0.01\textwidth][c]{R} &\makebox[0.01\textwidth][c]{C}\\
		\hline
		Spatial &78.3 &62.9 &\textbf{49.3} &\textbf{38.3} &\textbf{29.1} &\textbf{58.4} &123.9 &81.6 &66.5 &52.2 &40.2	&29.5 &59.7 &134.6\\
		Channel &78.5 &62.8 &49.0 &37.8 &29.0 &58.2 &123.8 &81.9 &\textbf{66.9} &52.5 &40.4	&29.5 &59.7 &134.9\\
		Parallel &\textbf{78.6} &\textbf{63.0} &49.1 &37.9 &28.9 &58.2 &123.1 &81.6 &66.7 &52.4 &40.3	&29.5 &59.7 &134.4\\
		Cascade &78.5 &62.8 &49.1 &38.0 &\textbf{29.1} &58.3 &\textbf{124.2} &\textbf{81.9} &66.8 &\textbf{52.6} &\textbf{40.6} &\textbf{29.6} &\textbf{59.9} &\textbf{135.4}\\
		\midrule
	\end{tabular*}
	\label{tab_7}
	\vspace{-3mm}
\end{table*}
\textbf{Evaluation on joint DMSE and DDR modules.} We incorporate the DMSE module and the DDR modules with four kinds of non-local interaction ways into the baseline, as shown in the 7-th to 10-th rows in Table~\ref{tab_2}. Clearly, our OSIC with both DMSE and DDR can deliver better results. Especially, cascading the non-local interactions in spatial and channel dimensions performs the best in most cases.

We also visualize the heatmaps of correlations between pixel features to analyze the effectiveness of DMSE and DDR in Figure~\ref{fig:heatmap_single}. Specifically, DMSE builds relatively explicit independence of features. DDR further refines the relationship between each feature point globally. For example, for the local features in the top left corner in Figure~\ref{fig:heatmap_single}, DMSE supplements and enriches the connections that are originally missing, and also weakens the unnecessary connections. Based on DMSE, the DDR module further refines the embedded features. That is, important connections are strengthened and unimportant connections are weakened, without changing the overall feature distributions.

From the above ablation studies, we conclude that both the DMSE and DDR modules are important to ensure the effectiveness of our OSIC for image captioning.

\textbf{Impact of the number of layers in the DDR module} We ablate our OSIC model with different configurations on the number of refining layers and modes of non-local modeling, as shown in Tables~\ref{tab_3},~\ref{tab_4},~\ref{tab_5} and~\ref{tab_6}, respectively. We vary the number of refining layers from $0$ to $6$. As the number of layers in the refining module is set to $0$, the model is greatly degraded. We also see that the setting with $4$ refining layers generally performs the best under the spatial non-local mode; the setting with $5$ refining layers generally outperforms other cases under the channel non-local mode. While for the parallel and cascade non-local mode, only $1$ refining layer can achieve the greatest improvement. That is, OSIC works better without lots of learnable parameters, which benefits the fast inferring and text generating.

\begin{figure}[h]
	\makeatletter
	\makeatother
	\centering
	\includegraphics[width=\linewidth]{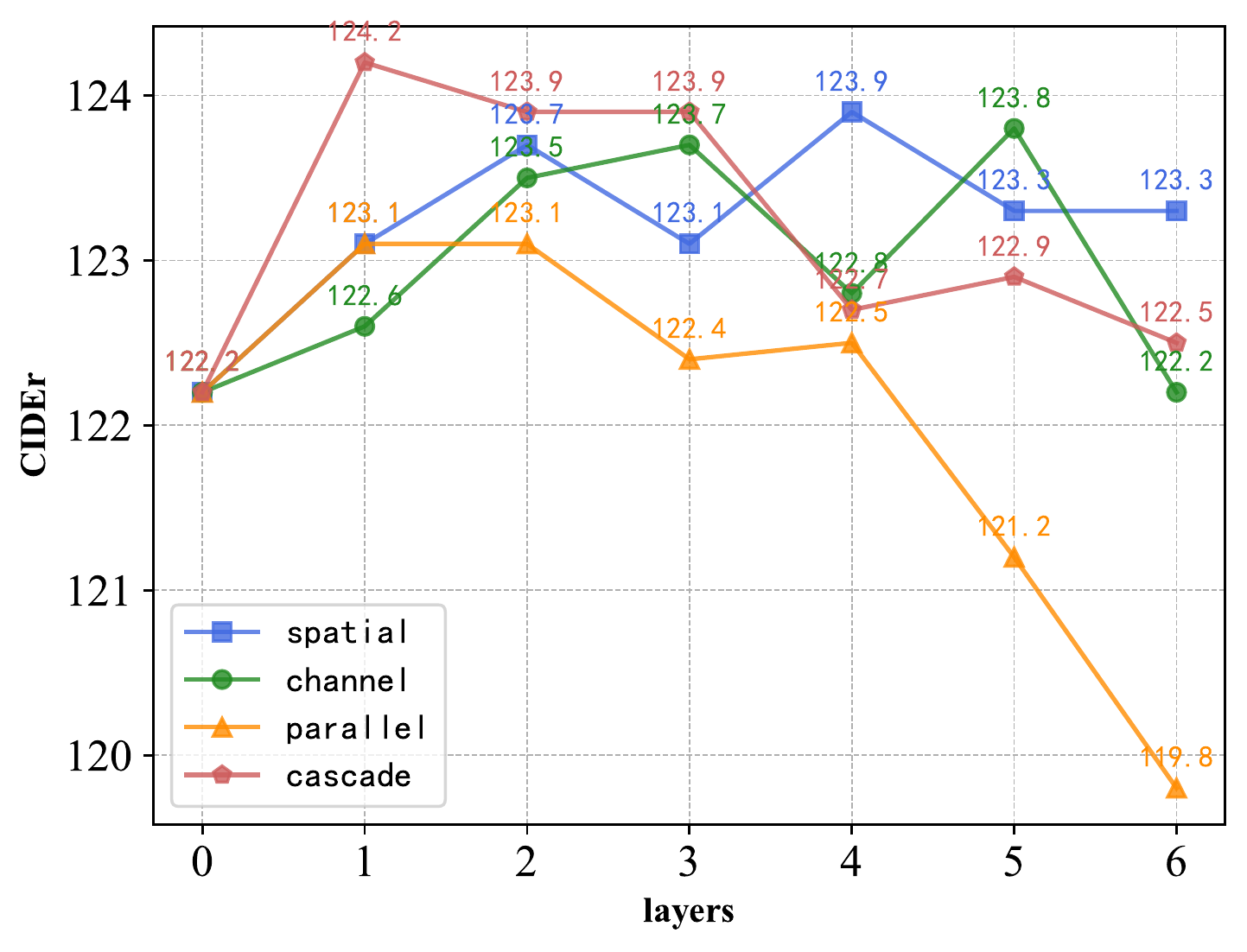}
	\vspace{-6mm}
	\caption{Evaluation of different refining modes under various numbers of layers. Where all the models are trained with the cross-entropy loss on Karpath’s split of MSCOCO dataset.}
	\label{fig:line_chart}
	\vspace{-6mm}
\end{figure}

\textbf{2) Further discussion on the best choices of the refining modes in terms of the CIDEr score.} We further explore the effects of different refining modes based on their best choices in Figure~\ref{fig:line_chart}. We see that the parallel refining mode generally performs the worst among different modes. It can also be seen that our OSIC model with only 1 layer cascade refining mode outperforms other cases. That is, cascading the spatial and channel dimensions is the best setting, which is used in all the experiments of this paper. As shown in Table~\ref{tab_7} to control sequence, our OSIC method with the cascaded spatial and channels still performs better than the other settings in the optimization of self-critical (SC) training. The evaluation on the models optimized by using SC training denotes that our OSIC with cascade dual-dimensional refining obtains the best performance.
\section{Conclusion}
We first defined the task-based information gap existed in current two-stage captioners, and addressed it by presenting a novel one-stage image captioner caleld OSIC. OSIC directly captures the global structure and local texture of input image by a new dynamic multi-sight learning encoder, and at the same time decodes the features into text descriptions in one stage. The features are generated and refined through building non-locally dual-dimensional information interaction to improve the global representation of the encoder for the image caption task. Extensive simulations demonstrated the effectiveness of our new one-stage image captioner, which is attributed to the dynamic multi-sight embedding and dual-dimensional refining modules, in comparison to current state-of-the-art methods. We also conduct extensive ablation studies to explore the importance of modules and the settings in each module, so that the best choice is chosen for experiments. In future, we will investigate more efficient interaction mechanism for one-stage image caption. Besides, the image caption tasks in complex conditions, such as describing images in rainy days~\cite{Wei2021DerainCycleGANRA, Wei2022SGINetTS} or blur scenes~\cite{Zhao2022FCLGANAL, Zhao2022CRNetUC}, are also worth investigating.

\section*{Acknowledgment}
This work is partially supported by the National Natural Science Foundation of China (62072151, 61932009, 61822701, 62036010, 72004174), and the Anhui Provincial Natural Science Fund for Distinguished Young Scholars (2008085J30). Zhao Zhang is the corresponding author of this paper. 

{\small
\bibliographystyle{ieee_fullname}
\bibliography{reference}
}

\end{document}